\documentclass[10pt,,compsoc]{IEEEtran}

\usepackage{amsmath}
\usepackage{mathtools}
\usepackage{graphics,epsfig}
\usepackage{xspace}
\usepackage{algorithm}
\usepackage{algorithmic}
\usepackage{amssymb}
\usepackage{subfigure}
\usepackage{ragged2e}
\usepackage{multirow}
\usepackage{multicol}
\usepackage{lineno,hyperref}

\begin{document}

\title{Fast $k$-means based on KNN Graph}

\author{Cheng-Hao Deng,
        ~and~Wan-Lei~Zhao
\IEEEcompsocitemizethanks{\IEEEcompsocthanksitem Fujian Key Laboratory of Sensing and Computing for Smart City, and the School of Information Science and Engineering, Xiamen University, Xiamen, 361005, P. R. China.\protect\\
Wan-Lei Zhao is the corresponding author. E-mail: wlzhao@xmu.edu.cn. }}

\IEEEtitleabstractindextext{
\begin{justify}
\begin{abstract}
In the era of big data, $k$-means clustering has been widely adopted as a basic processing tool in various contexts. However, its computational cost could be prohibitively high as the data size and the cluster number are large. It is well known that the processing bottleneck of $k$-means lies in the operation of seeking closest centroid in each iteration. In this paper, a novel solution towards the scalability issue of $k$-means is presented. In the proposal, $k$-means is supported by an approximate k-nearest neighbors graph. In the $k$-means iteration, each data sample is only compared to clusters that its nearest neighbors reside. Since the number of nearest neighbors we consider is much less than k, the processing cost in this step becomes minor and irrelevant to k. The processing bottleneck is therefore overcome. The most interesting thing is that k-nearest neighbor graph is constructed by iteratively calling the fast $k$-means itself. Comparing with existing fast $k$-means variants, the proposed algorithm achieves hundreds to thousands times speed-up while maintaining high clustering quality. As it is tested on 10 million 512-dimensional data, it takes only 5.2 hours to produce 1 million clusters. In contrast, to fulfill the same scale of clustering, it would take 3 years for traditional $k$-means. 

\end{abstract}
\end{justify}

}

\maketitle

\section{Introduction}
\label{sec:intro}
Clustering problems arise from a wide variety of applications such as knowledge discovery~\cite{ml04:zhao}, data compression~\cite{JPDSPS11}, large-scale image linking~\cite{webhashclustering} and visual vocabulary construction~\cite{SiZ03}. Since the general $k$-means algorithm~\cite{km82, kmeans} was proposed in 1982, continuous efforts have been made to search for better solution for this issue. Various algorithms have been proposed in the last two decades, such as mean shift~\cite{meansift}, DB-SCAN~\cite{dbscan}, spectral clustering~\cite{spectral}, Rank-Order~\cite{rankorder} BIRCH~\cite{birch1996} and Clusterdp~\cite{science14}, etc. Among these algorithms, \textit{k}-means \cite{kmeans} remains popular for its simplicity, efficiency and moderate but stable performance under different contexts. It is known as one of top ten most popular algorithms in data mining~\cite{top10}. 

In traditional $k$-means, given a set of $n$ data samples in real \textit{d}-dimensional space $R^d$, and an integer \textit{k}, clustering is modeled as a distortion minimization process. The clustering process partitions $n$ samples into $k$ sets such that to minimize the mean squared distance from each sample to its nearest cluster centroid. It could be formularized as 

\begin{equation}
        \mbox{min }\sum_{q(x_i)=r}{\parallel C_r -x_i \parallel^2},
        \label{eqn:tkm}
\end{equation}
where $x_i \in R^d$ and $C_r$ is the centroid of cluster $r$. In Eqn.~\ref{eqn:tkm}, function $q(x_i)$ returns the closest centroid (among k centroids) for sample $x_i$. In general, there are two major steps in the \textit{k}-means iteration. In the first step, each sample is assigned to its closest centroid. In its second step, each centroid $C_r$ is updated by taking the average over assigned samples. The above two steps are repeated until there is no distortion change (Eqn.~\ref{eqn:tkm}) in two consecutive iterations.

Although $k$-means remains popular, it actually suffers from two major issues. Firstly, it is well-known that $k$-means only converges to local optima. Recent researches have been working on improving its clustering quality~\cite{kpp07,rbkmeans,boostkmeans}. Thanks to the introduction of incremental optimization strategy in~\cite{boostkmeans}, $k$-means is able to converge to considerably lower distortion. 

The second issue is mainly about its scalability. Although, the complexity of $k$-means is linear to the size of input data, the clustering cost could be prohibitively high given both the size of data and the expected number of clusters $k$ are very large. Moreover, according to~\cite{ailon09,vattani11}, in its worst case, the running time for $k$-means could be exponential against the size of input samples. Due to the steady growth of data volume in various forms (web-pages, images, videos, audios and business transactions) on a daily basis, the scalability issue of this traditional algorithm becomes more and more imminent. In each $k$-means iteration, the most intensive operation is of assigning samples to their closest centroid. As a result, the scalability issue principally is due to the heavy cost of computing nearest centroid for each sample, which is $O(n{\cdot}d{\cdot}k)$.

In recent years, continuous efforts have been devoted to looking for effective solutions that are still workable in web-scale data. Representative works are~\cite{efficient,mnkm10, kpp12,akm,ikm,drvq,egm,ikmn15,cckm12, nipfast}. However, most of the $k$-means variants achieve high speed efficiency while sacrificing the clustering quality. Algorithm presented in~\cite{icml03:elkan} demonstrates faster speed and maintains relatively high quality. Unfortunately, a lot of extra memory are required. Specifically, its memory complexity is quadratic to \textit{k}, which turns out to be unsuitable in the case that \textit{k} is very large.

In this paper, an efficient $k$-means variant is proposed, in which the $k$-means clustering process is supported by an approximate k-nearest neighbor graph (KNN graph). The approximate KNN graph is built in its pre-processing step, in which the fast $k$-means itself and a KNN graph construction process are jointly undertaken. We interestingly discover that, these two processes could be actually beneficial to each other. This idea is inspired by the following observation
\begin{itemize}
	\item {With high probability that one sample and its nearest neighbors reside in the same cluster}
\end{itemize}

\begin{figure}
\begin{center}
  \subfigure[\textit{k}-means]
  {\includegraphics[width=0.485\linewidth]{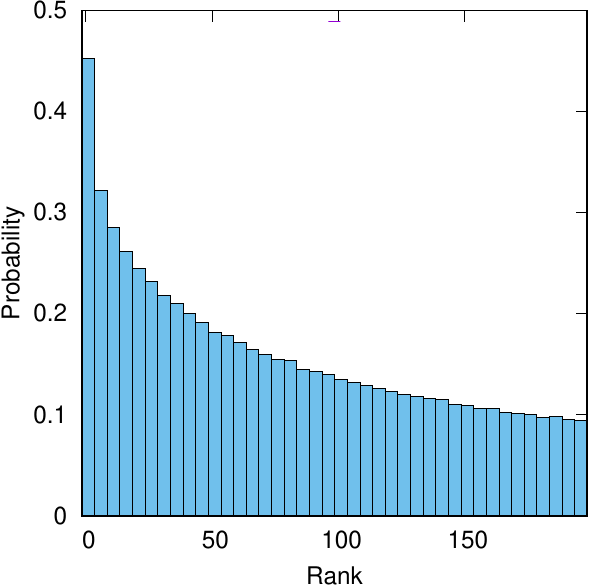}}
  \hspace{0.01in}
  \subfigure[2M tree]
  {\includegraphics[width=0.485\linewidth]{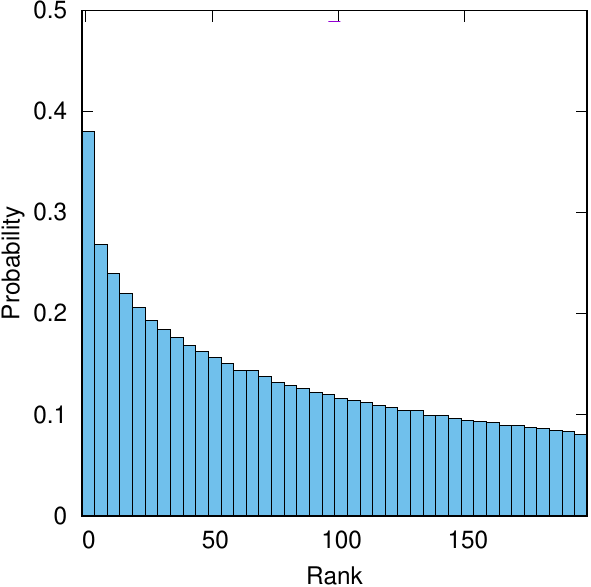}}
  \caption{The stastitics on co-occurrence rate of one sample and its \textit{k}-th nearest neighbor that in the same cluster. Experiments have been conducted on SIFT100K~\cite{pq} with traditional $k$-means~\cite{km82} and two-means tree~\cite{2mtree}. In the experiment, the size of each cluster is fixed to \textit{50}. Note that the probability of two randomly selected samples that fall into the same cluster is only $50/100000 = 0.0005 \ll 0.1$.}
\label{fig:mov}
\end{center}
\end{figure}

Fig.~\ref{fig:mov} shows the co-occurrence rate of one sample and its $\kappa$-th nearest neighbor in one cluster. \textit{k}-means and its variants two-means tree~\cite{2mtree} are tested on SIFT100K~\cite{pq}. The same trend is observed in both cases. If the samples are closer, the probability that they appear in the same cluster is higher. This probability is much higher than the probability of a random collision.

With this observation, we learn that one sample and its nearest neighbors should be arranged into the same cluster. On one hand, this indicates if one sample and its neighbors temporarily are not in the same cluster, it is reasonable to compare one sample only to the clusters where its neighbors reside. Among these clusters, there is probably a true one that all the neighboring samples should live together. On the other hand, from the viewpoint of k-nearest neighbor graph (KNN graph) construction, in order to build KNN list for one sample, it is sufficient to compare one sample to samples reside in the same cluster since its neighbors are most likely reside in the same cluster. 

Based on the above analysis, the scalability issue of $k$-means clustering is addressed in two steps in the paper. Firstly, the fast $k$-means is called to build an approximate KNN graph for itself. Secondly, the fast $k$-means clustering is undertaken with the support of constructed KNN graph. During the fast $k$-means iteration, one sample will only compare to the clusters that its top-$\kappa$ nearest neighbors reside. Usually, the number of nearest neighbors we consider is considerably smaller than the clustering number \textit{k}. The number of clusters we actually visit is even fewer. As a consequence, significant speed-up is expected. Moreover, as revealed later, the clustering quality drops very little with such kind of speed-up scheme.

Although our primary goal is to speed-up the $k$-means clustering with the support of KNN graph, we interestingly discover that satisfactory performance is achieved when the constructed KNN graph is applied on the approximate nearest neighbor search (ANNS) tasks. Moreover, comparing with the KNN graph construction algorithms that are specifically designed for approximate nearest neighbor search~\cite{weidong, efenna16, hnsw16}, our algorithm requires much lower computational cost.

The remainder of this paper is organized as follows. The reviews about representative works on improving the performance of traditional $k$-means are presented in Section~\ref{sec:relat}. In Section~\ref{sec:pre}, two important $k$-means variants are reviewed, which facilitates the discussion of our algorithm in Section~\ref{sec:alg}. Extensive experiment studies over proposed clustering method are presented in Section~\ref{sec:exp}. Section~\ref{sec:conl} concludes the paper.

\section{Related Works}
\label{sec:relat}
\subsection{\textit{k}-means Variants}
Due to its versatility and simplicity, $k$-means has been widely adopted in different contexts. In the era of big-data, $k$-means has been used as a basic tool to process large-scale data of various forms. Unfortunately, as discussed in Section~\ref{sec:intro}, the computational cost could be prohibitively high as the scale of data increases to extraordinarily large, i.e. billion level. Recently, several $k$-means variants are proposed to either enhance its clustering quality or scalability.

In terms of the clustering quality, one of the important work comes from S. Vassilvitskii et al.~\cite{kpp07,kpp12}. The motivation is based on the observation that $k$-means converges to a better local optima if the initial clustering centroids are carefully selected. According to~\cite{kpp07}, $k$-means iteration also converges faster due to the careful selection on the initial cluster centroids. However, in order to adapt the initial centroids to the data distribution, \textit{k} rounds of scanning over the data are necessary. Although the number of scanning rounds has been reduced to a few in~\cite{kpp12}, the extra computational cost is still inevitable.

Recently, a new variant called boost $k$-means is proposed~\cite{boostkmeans}. The ``egg-chicken'' loop in $k$-means has been simplified as a stochastic optimization process, which is also known as incremental $k$-means~\cite{ml04:zhao}. As indicated by extensive experiments, it is able to converge to a considerably better local optima while involving no extra cost. Due to its superior performance, this incremental optimization scheme is adopted in our design. In order to facilitate our discussion, a more detailed review about boost $k$-means is given in Section~\ref{sec:pre}.

In each $k$-means iteration, the processing bottleneck is the operation of assigning each sample to its closest centroid. The iteration becomes unbearably slow when both the size and the dimension of the data are very large. Noticed that this is a nearest neighbor search problem, Kanungo et al.~\cite{dmount02} proposed to index dataset in a KD Tree~\cite{kdtree75} to speed-up the sample-to-centroid nearest neighbor search. Unfortunately, this is only feasible when the dimension of data is in few tens. Similar scheme has been adopted by Dan et al.~\cite{pelleg99}. However, due to the curse of dimensionality, this method becomes ineffective when the dimension of data grows to a few hundreds. A recent work~\cite{wsdm14} takes similar way to speed-up the nearest neighbor search by indexing dataset with inverted file structure. During the iteration, each centroid is queried against all the indexed data. Attributing to the efficiency of inverted file structure, one to two orders of magnitude speed-up is observed. However, inverted file indexing structure is only effective for sparse vectors.

Alternatively, the scalability issue of $k$-means is addressed by subsampling over the dataset during $k$-means iteration. Namely, methods in~\cite{mnkm10,icdm04} only pick a small portion of the whole dataset to update the clustering centroids each time. For the sake of speed efficiency, the number of iterations is empirically set to small value. It is therefore possible that the clustering terminates without a single pass over the whole dataset, which leads to higher speed but also higher clustering distortion. Even though, when coping with high dimensional data in big size, the speed-up achieved by these methods are still limited.

Apart from above methods, the speed-up could be achieved by reducing the comparisons between samples and centroids.
In~\cite{cckm12}, only the ``active points'', which are the samples located on the cluster boundaries, are considered to be swapped between clusters. Comparing with other $k$-means variants, it makes a good trade-off between efficiency and clustering quality whereas considerable quality degradation is still inevitable.

Another easy way to reduce the number of comparisons between samples and centroids is to conduct the clustering in a top-down hierarchical manner~\cite{ml04:zhao,jain88, kddzhao05}. Specifically, the clustering solution is obtained via a sequence of repeated bisections. The clustering complexity of $k$-means is reduced from $O(t{\cdot}k{\cdot}n{\cdot}d)$ to $O(t{\cdot}log(k){\cdot}n{\cdot}d)$~\cite{boostkmeans}. This is particularly significant when \textit{n}, \textit{d} and \textit{k} are all very large. However, poor clustering performance is achieved in the usual case as it breaks the \textit{Lloyd}'s condition~\cite{boostkmeans}.

\subsection{K-Nearest Neighbor Graph Construction}
KNN graph is primarily built to support nearest neighbor search~\cite{weidong,jmlr09}. It is also the key data structure in the manifold learning and machine learning, etc~\cite{weidong}. Basically, it tries to find the top-$\kappa$ nearest neighbors for each data point. When it is built in brute-force way, its time complexity is $O(d{\cdot}n^2)$, where both $d$ and $n$ could be very large. As a result, it is computationally expensive to build an exact KNN graph. For this reason, recent works~\cite{weidong,efenna16, jmlr09, msraKNN} aim to search for an approximate but efficient solution. In~\cite{jmlr09}, an approximate KNN graph is built efficiently by divide-and-conquer strategy. In this algorithm, the original dataset is partitioned into thousands of small subsets by KD trees. KNN list is built by exhaustive comparison within each subset. However, the recall of KNN graph turns out to be very low. Recent works~\cite{efenna16, msraKNN} could be viewed as improvements over this work. In 2011, a very successful KNN graph construction algorithm called NN Descent/KGraph \cite{weidong} has been proposed. This algorithm is proposed based on the observation that ``a neighbor of a neighbor is also likely to be a neighbor''. According to~\cite{weidong}, its empirical time complexity is only $O(n^{1.14})$. Unfortunately, according to our observation, its recall drops dramatically as the scale of data increases to very large, i.e. 10M. Algorithm presented in~\cite{efenna16} faces similar problem.

In this paper, a novel KNN graph construction algorithm is proposed and used to support the fast $k$-means clustering. To the best of our knowledge, this is the first piece of work that KNN graph is used to speed-up $k$-means clustering. In addition, comparing with other KNN graph construction algorithms, our algorithm is computationally efficient and leads to lowest clustering distortion. Furthermore, when it is applied in ANNS problem, it shows satisfactory performance across different datasets.

%NN Descent is very efficient in SIFT1M dataset, but it very hard to get a high recall in VLAD10M dataset due to its large scale and high dimension (This two datasets will be introduced carefully in \ref{sec:experiment}). 

%We have test $k$-means clustering above best KNN Graph, but the clustering result inferior to our iteration $k$-means KNN Graph based $k$-means.

\section{PRELIMINARIES}
In order to facilitate our discussions in later sections, two important $k$-means variants are reviewed, namely, boost $k$-means (BKM)~\cite{boostkmeans} and two means (2M) tree~\cite{2mtree}. As shown later, our speed-up scheme is built upon boost $k$-means instead of traditional $k$-means as the former always produces clusters of higher quality. While two means tree is used to produce initial clusters for its high efficiency.
\label{sec:pre}
\subsection{Boost \textit{k}-means}
As an extension of incremental $k$-means~\cite{ml04:zhao}, boost \textit{k}-means allows the optimization iteration to be feasible for the whole $\textit{l}_2$ space. Different from other $k$-means variants, boost $k$-means iteration is driven by an explicit objective function. Given clusters $S_{r=1{\cdots}k}$, the composite vector of a cluster is defined as $D_r=\sum_{x_i \in S_r}x_i$. The objective function of boost $k$-means is written as
\begin{equation}
          \mathcal{I} = \sum_{r=1}^k\frac{D_r'D_r}{n_r},
         \label{eqn:iz}
\end{equation}
which is directly derived from Eqn.~\ref{eqn:tkm}. With this objective function, the traditional $k$-means clustering is revised to a stochastic optimization procedure. Each time, one sample is randomly selected and searches for a better re-allocation that leads to highest increase of $\mathcal{I}$. Namely, the variation of function value that is incurred by the possible movement (moving $x_i$ from $S_u$ to $S_v$) is given by
\begin{equation}
\begin{aligned}
   {\Delta}\mathcal{I}(x_i)=&\frac{(D_v+x_i)'(D_v+x_i)}{n_v+1}+\frac{(D_u-x_i)'(D_u-x_i)}{n_u-1}  \\
   &- \frac{D_v'D_v}{n_v}-\frac{D_u'D_u}{n_u}.
\end{aligned}
\label{eqn:val}
\end{equation}

The optimization process seeks for the movement that ${\Delta}\mathcal{I}(x_i)$ is the highest and positive. In particular, the movement of $x_i$ from $S_u$ to $S_v$ is made as soon as we find the movement is appropriate. According to~\cite{boostkmeans}, it is able to converge to a much better local optima in comparison to $k$-means and its variants. The cost of checking the best movement in boost $k$-means is equivalent to seeking for closest centroid in traditional $k$-means. As a result, boost $k$-means is on the same complexity level as traditional $k$-means. 

Due to its superior performance, boost $k$-means is fully adopted in our design. Particularly, the speed-up we made in this paper is on boost $k$-means instead of traditional $k$-means.
 
\subsection{Two Means Tree}
Two means (2M) tree~\cite{2mtree} is a variant of hierarchical bisecting $k$-means. It has been adopted in KNN graph construction for its high speed efficiency~\cite{2mtree}. Alg.~\ref{alg:2m} shows the general procedure of two means tree. Similar as bisecting $k$-means, the samples are partitioned recursively into two clusters each time until \textit{k} clusters are produced. Different from bisecting $k$-means, one more step is taken in the end of each bisecting. The resulting two clusters are adjusted to equal size. Its complexity is the same as bisecting $k$-means, namely $O(d{\cdot}n{\cdot}log(n))$, which is even faster than one round $k$-means iteration. In this paper, two means tree is adopted only to generate initial $k$-means partition. In order to enhance its performance, the aforementioned boost $k$-means is integrated in the bisecting operation (\textit{Step 8} in Alg.~\ref{alg:2m}).

\begin{algorithm}{\textbf{TwoMeans}($X_{n{\times}d}$, $k$)}
 \begin{algorithmic}[1]
  \STATE \textbf{Input}: matrix $X_{n{\times}d}$
  \STATE \textbf{Output}: matrix cLabel$_{n{\times}1}$
  \STATE $t = 1$. 
  \STATE cLabel[$1{\cdots}n$] $\leftarrow 1$;
  \STATE Map cLabel[${1{\cdots}n}$] to partition $\mathcal{S}$;
  \WHILE {$t < k$}
  	\STATE Pop $S_i$ with largest size out of $\mathcal{S}$
  	\STATE Bisect $S_i$ into $S_{u}$ and $S_v$
  	\STATE Adjust $S_{u}$ and $S_v$ to equal size
  	\STATE $\mathcal{S} \leftarrow \mathcal{S} \cup S_u \cup S_v$
	\STATE $t = t + 1$;
  \ENDWHILE
  \STATE Map $\mathcal{S}$ to cLabel[${1{\cdots}n}$];
 \end{algorithmic}
 \label{alg:2m}
\end{algorithm}

In order to facilitate the operations in later steps, the mapping at \textit{Line 5} converts cluster labels of samples into cluster set $\mathcal{S}$. At the end of two means tree clustering, cluster set $\mathcal{S}$ is mapped back as cluster label representation at \textit{Line 13}.

\section{KNN Graph based \textit{k}-means}
\label{sec:alg}
In this section, our solution to the scalability issue of $k$-means is presented. Firstly, a general procedure that how boost $k$-means is undertaken with the support of KNN graph is given. To support fast clustering, the process of KNN graph construction should be sufficiently fast otherwise it becomes another processing bottleneck. To overcome this problem, a novel light-weight KNN graph construction procedure is also introduced.

\subsection{Motivation}
As illustrated in Fig.~\ref{fig:motivation}, there is a strong correlation between the closeness of data samples and their membership living in one cluster. This correlation could be interpreted from either the side of $k$-means clustering or the side of KNN graph construction.

\begin{itemize}
	\item {From the clustering side, if the KNN list of each sample is known, clustering is a process of arranging close neighbors into one cluster. As a result, given one sample, the clustering only needs to check with the clusters its $\kappa$-nearest neighbors live in. Such that it seeks the approriate cluster to move to. It is therefore no need to check with all $k-1$ clusters. As a consequence, the processing bottleneck is overcomed.}
	\item {From the KNN graph construction side, if the data samples are already partitioned into small clusters, KNN graph construction is undertaken within each cluster by an exhaustive pair-wise comparison. As a consequence, the KNN graph construction is pulled out in a very efficient manner.}
\end{itemize}

Based on the first piece of interpretation, we work out the fast $k$-means algorithm. Similarly, based on the second piece of interpretation, KNN graph construction algorithm is conceived.

\subsection{Fast \textit{k}-means Driven by KNN Graph}
Given a KNN graph is ready, boost $k$-means procedure presented in~\cite{boostkmeans} is revised as Alg.~\ref{alg:kmeans}. At the beginning of the clustering, 2M tree (Alg.~\ref{alg:2m}) is called to produce \textit{k} clusters. The initial clusters will be incrementally optimized in the later steps. In each step of the optimization iteration, one sample is randomly selected. Thereafter, all the clusters in which its $\kappa$ neighbors reside are collected. The selected sample is therefore checked with these clusters to  seek for the best move. The iteration terminates until convergence condition is reached.

\begin{algorithm}{\textbf{GK-means}($X_{n{\times}d}$, \textit{k}, $G_{n{\times}\kappa}$)}
 \begin{algorithmic}[1]
  \STATE \textbf{Input}: matrix $X_{n{\times}d}$, \textit{k}, KNN graph $G_{n{\times}\kappa}$
  \STATE \textbf{Output}: $S_1,{\cdots},S_r,{\cdots}S_k$
  \STATE cLabel = \textbf{TwoMeans}($X_{n{\times}d}$, k);
  \STATE $Q{\leftarrow} \varnothing$;
  \WHILE {not convergence}
  \FOR {each $x_i \in X$}
  \FOR {j = 1; j $\leq \kappa$;}
  	\STATE b  = G[i][j];
    \STATE $Q\leftarrow Q \cup~$cLabel[b];
    \STATE j  = j + 1;
  \ENDFOR
  \STATE Seek $v$ in Q that maximizes ${\Delta}\mathcal{I}(x_i)$;
  \IF {${\Delta}\mathcal{I}(x_i) > 0$}
	  \STATE Move $x_i$ from current cluster to $S_v$;
  \ENDIF
  \STATE $Q{\leftarrow} \varnothing$;
  \ENDFOR
  \ENDWHILE
 \end{algorithmic}
 \label{alg:kmeans}
\end{algorithm}

Comparing with the procedure presented in boost $k$-means~\cite{boostkmeans}, there are basically two major modifications. Firstly, the initial clusters are initialized by two means tree, whose complexity is only $O(n{\cdot}log(k){\cdot}d)$~\cite{boostkmeans}. It is  considerably faster than traditional $k$-means initialization. Secondly, as shown from \textit{Line 6-12}, only clusters that keep the first $\kappa$ neighbors of $x_i$ are visited, the number of which is much smaller than \textit{k}. Furthermore, it is possible that several neighbors of $x_i$ may live in the same cluster. As a consequence, the number of clusters that one sample visits is even smaller than $\kappa$.

Alg.~\ref{alg:kmeans} is built upon boost $k$-means. Alternatively, similar speed-up is also feasible for traditional $k$-means. To achieve that, \textit{Line 12-15} in Alg.~\ref{alg:kmeans} is modified to seeking for the closest centroid from the collected clusters. In Section~\ref{sec:exp}, the performance of this alternative configuration will be presented. As will be revealed, similar speed-up is achieved whereas it shows inferior clustering quality in comparison to the one built upon boost $k$-means.

\subsection{KNN Graph Construction with Fast \textit{k}-means}
As discussed in Section~\ref{sec:intro}, samples live in the same cluster are likely to be neighbors. In the KNN graph construction,
this clue could be fully exploited. Namely, the search for $\kappa$ nearest neighbors for one sample is undertaken within the cluster it resides. Based on this principle, the fast KNN graph construction is conceived. Firstly, fast $k$-means clustering (\textit{Line 6}, Alg.~\ref{alg:kmeans}) is called to produce fixed number of clusters. Thereafter, exhaustive comparisons are conducted within each  cluster. The new closer point pairs are used to update the KNN graph (\textit{Line 7-13}, Alg.~\ref{alg:knng}).

\begin{figure}
\begin{center}
\includegraphics[width=0.70\linewidth]{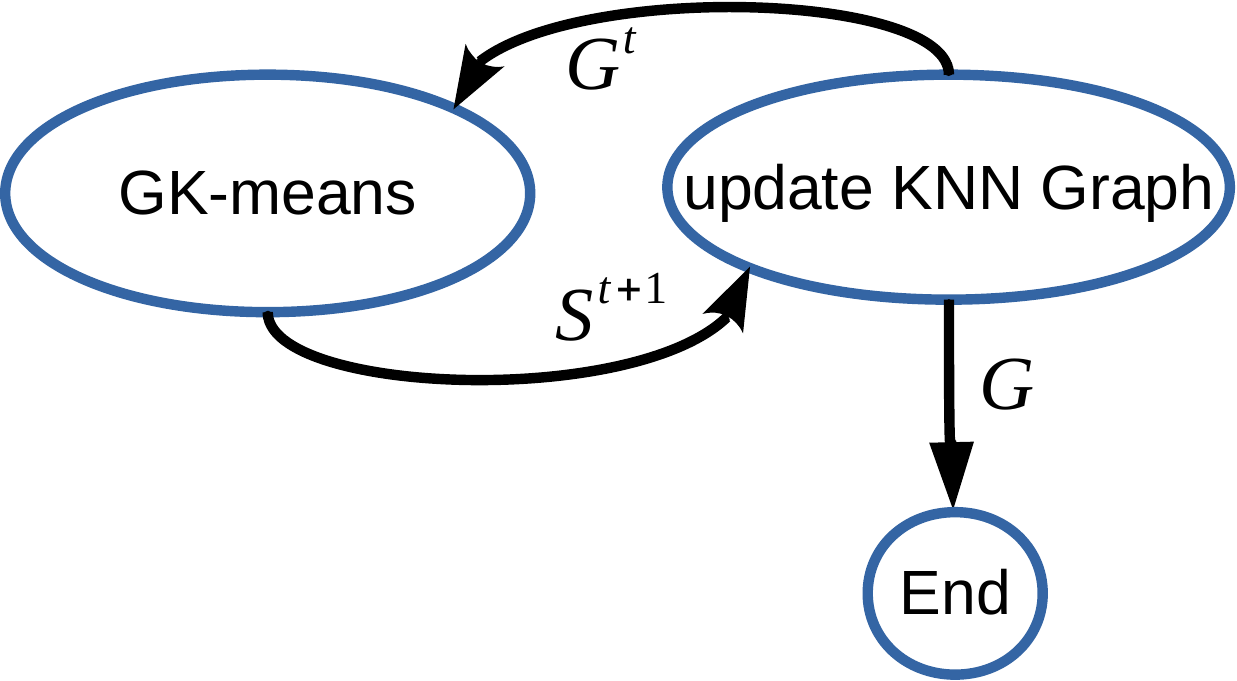}
\caption{The illustration of the intertwined evolving process in the KNN graph construction.}
\label{fig:gkm}
\end{center}
\end{figure}

In order to control the complexity of KNN graph construction on a low level, the cluster size is fixed to a small constant $\xi$, i.e. \textit{50}. Given the cluster size is fixed to a constant, it is easy to see the cluster number is $k_0 = {\lfloor}\frac{n}{\xi}{\rfloor}$. According to Alg.~\ref{alg:kmeans}, a KNN graph is required as an input parameter. In our design, a random KNN graph is supplied at the beginning. Since the KNN graph is randomly initialized, one would not expect good cluster partitions returned by Alg.~\ref{alg:kmeans} at the beginning. However, as the iteration continues, the quality of KNN graph $G^t$ is enhanced incrementally. Accordingly, the structure of cluster partitions returned by Alg.~\ref{alg:kmeans} becomes better. As a result, the structures of KNN graph and the cluster structures evolve alternatively. Fig.~\ref{fig:gkm} illustrates this intertwined evolving process. The iteration parameter $\tau$ controls the final quality of KNN graph. Larger $\tau$ leads to preciser KNN graph while taking higher time cost.

\begin{algorithm}{KNN Graph Construction}
  \begin{algorithmic}[1]
    \STATE \textbf{Input}: $X_{n{\times}d}$: reference set, $\kappa$: scale of k-NN Graph
    \STATE \textbf{Output}: KNN Graph $G_{n{\times}\kappa}$
    \STATE t $\leftarrow$ 0; 
    \STATE Initialize $G^t_{n{\times}\kappa}$ with random lists;
    \STATE $k_0 = {\lfloor}\frac{n}{\xi}{\rfloor}$;
    \FOR {t $<$ $\tau$}
    	\STATE $\mathcal{S}$ = \textbf{GK-means}($X_{n{\times}d}$, $k_0$, $G^t$)
	    \FOR {each $S_m \in \mathcal{S}$}
		    \FOR {each $<i, j>_{(i < j)} \in S_m {\times}S_m$}
  			    \IF { $<i, j>$ is NOT visited}
				    \STATE Update $G^{t}[i]$ and $G^{t}[j]$ with $d(x_i, x_j)$;
				 \ENDIF
		    \ENDFOR
		\ENDFOR
		\STATE t $\leftarrow$ t + 1
    \ENDFOR
    \STATE $G \leftarrow G^t$;
  \end{algorithmic}
  \label{alg:knng}
\end{algorithm}

\begin{figure}
\begin{center}
\includegraphics[width=0.70\linewidth]{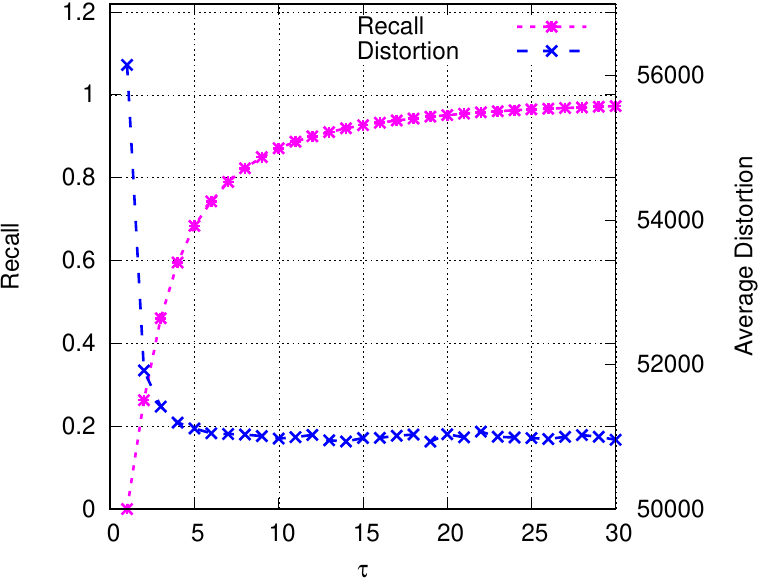}
\caption{The correlation between the KNN graph recall and clustering distortion shown as the function of $\tau$. The experiment is  conducted on SIFT100K~\cite{pq}.}
\label{fig:motivation}
\end{center}
\end{figure}

Fig.~\ref{fig:motivation} shows the curves of average recall (top-1) of KNN graph and clustering distortion as the functions of $\tau$. As shown in the figure, at the beginning of this procedure, both the quality of clustering and the quality of KNN graph are very poor. The clustering results are nearly random. Correspondingly, the average recall of KNN graph is close to \textit{0}. However, after only \textit{5} iterations, the clustering distortion drops considerably. In the meantime, the average recall increases to above \textit{0.6}.

In Alg.~\ref{alg:kmeans}, there is no specification that which KNN graph construction algorithm should be adopted. As a result, KNN graph supplied by any construction algorithms will achieve similar speed-up. However, as revealed later, the KNN graph algorithm presented in Alg.~\ref{alg:knng} produces the best clustering quality. Moreover, comparing with other KNN graph construction algorithms~\cite{weidong, hnsw16, efenna16}, Alg.~\ref{alg:knng} takes much less memory. The extra memory it takes is to keep the KNN graph. Furthermore, it is at least two times faster than NN Descent~\cite{weidong} and small world graph construction~\cite{hnsw16}. 

Due to its low computational cost, Alg.~\ref{alg:knng} can be also adopted to construct KNN graph for approximate nearest neighbor search. According to our observation, although the quality of KNN graph (measured by recall) is usually lower than that of NN Descent~\cite{weidong}, it is able to achieve similar or even better performance than the methods presented in~\cite{hnsw16,nndecent}. For instance, it takes less than \textit{3}ms to fulfill a query on \textit{100} million SIFTs~\cite{pq} with its recall above \textit{0.9}. The full discussion about ANNS with our KNN graph is beyond the focus of this paper.

As a summary, the proposed fast $k$-means consists of two major steps. In the first step, the fast $k$-means is called to build an approximate KNN graph for itself. In the second step, the fast $k$-means is performed again to produce \textit{k} clusters with the support of the approximate KNN graph. Since the KNN graph is built based on the intermediate clustering results in the first step, the information that how the samples are organized as clusters is kept with the KNN graph. The clustering in the second step is therefore guided by such kind of ``prior knowledge''. Since this algorithm is based on KNN graph, it is called as graph based $k$-means (GK-means) from now on.

\subsection{Discussion on Parameters}
In Alg.~\ref{alg:kmeans} and Alg.~\ref{alg:knng}, besides the cluster number \textit{k}, there are additionally three parameters are involved. Parameter $\tau$ in Alg.~\ref{alg:kmeans} controls the quality of KNN graph. According to our observation, it is sufficient to set $\tau=10$ for clustering task. While if Alg.~\ref{alg:knng} is called to produce KNN graph for ANNS task, $\tau=10$ could be set up-to, i.e. \textit{32}. Parameter $\xi$ controls the size of cluster that is used for KNN graph construction. Larger $\xi$ leads to better KNN graph quality whereas it also induces more number of pair-wise comparisons. For this reason, a trade-off has to be made. According to our observation, the recommended range of $\xi$ is $[40, 100]$. Parameter $\kappa$ controls the number of neighbors that one sample should consider during the fast $k$-means clustering. This in turn determines the number of clusters that one sample visits. If only few neighbors are considered, the chance that we miss the true cluster will be high. On the other hand, if too many neighbors are considered during the comparison, a lot of comparisons are required. The speed-up over traditional $k$-means becomes less significant. Again a trade-off has to be made. According to our empirical study, the clustering quality becomes very stable as $\kappa$ is larger than \textit{40}. In our implementation, $\tau$, $\xi$ and $\kappa$ are fixed to \textit{10}, \textit{50} and \textit{50} respectively.

\subsection{Complexity Analysis}
In this section, the complexity of Alg.~\ref{alg:kmeans} and Alg.~\ref{alg:knng} is analyzed. As shown above, GK-means (Alg.~\ref{alg:kmeans}) is comprised by two major parts, namely two means initialization and fast $k$-means clustering. For the first part, the complexity of 2M tree initialization is $O(d{\cdot}n{\cdot}log(k))$~\cite{boostkmeans}. For the second part, since one sample only visits at most $\kappa$ clusters in the iteration, the cost of clustering is only $d{\cdot}n{\cdot}\kappa$ in each iteration. As a result, the overall complexity is $O(d{\cdot}n{\cdot}log(k)+{t}{\cdot}d{\cdot}n{\cdot}\kappa)$, where $t$ is the number of iterations. From above analysis, it is clear to see that the cluster number \textit{k} has very minor impact on the clustering complexity.

The KNN graph construcion (Alg.~\ref{alg:knng}) consists of two major steps, namely fast $k$-means clustering and KNN graph refinement. In the clustering step, according to above analysis, its complexity is $O(d{\cdot}n{\cdot}log(\frac{n}{\xi})+d{\cdot}n{\cdot}\kappa)$. Noticed that, $t$ is fixed to \textit{1} in the KNN graph construction. In KNN graph refinement step, one sample is compared to around $\xi$ samples. Therefore, its complexity is $O(d{\cdot}n{\cdot}\xi)$. As a result, the complexity of KNN graph construction is $O(d{\cdot}n{\cdot}log(\frac{n}{\xi})+{\cdot}d{\cdot}n{\cdot}\kappa+d{\cdot}n{\cdot}\xi)$, where both $\xi$ and $\kappa$ are small constants. Overall, the complexity of the whole procedure is on $O(d{\cdot}n{\cdot}log(n))$ level.

\section{Experiments on Clustering Task}
\label{sec:exp}

In this section, the performance of GK-means is studied in comparison to $k$-means and its representative variants such as boost $k$-means (BKM)~\cite{boostkmeans}, closure $k$-means~\cite{cckm12} and Mini-Batch~\cite{mnkm10}. AKM~\cite{akm} and HKM~\cite{pami14:flann} are not considered as inferior performance to closure $k$-means is reported in~\cite{cckm12} . Before the comparisons with other $k$-means variants, the performance of GK-means is studied under different configurations. Namely, we try to see how well GK-means is performed when it is built upon traditional $k$-means instead of boost $k$-means. Additionally, GK-means is also tested when the approximate KNN graph is supplied by NN Descent~\cite{weidong}. To do that, we want to search for the best configuration that we can currently set for GK-means.

We mainly study the clustering quality and scalability on four large-scale datasets, which are summarized in Tab.~\ref{tab:datasets}. The type of data covers from image local features, image global features to vectorized text word features. The dimension of data varies from \textit{100} to \textit{960} dimensions. The scale of all the datasets are above \textit{1} million level. All the methods considered in the paper are implemented in C++ and compiled with GCC \textit{5.4}. The simulations are conducted by single thread on a PC with \textit{2.4}GHz Xeon CPU and \textit{32}G memory setup.

\begin{table}
\begin{center}
\caption{Overview of Datasets}
\vspace{-0.1in}
\footnotesize{
\begin{tabular}{|l|c|c|c|c|l|l|}
\hline
\multirow{2}{*} {Datasets} & \multicolumn{2}{c|}{Scale} & \multirow{2}{*} {Data Type}  \\
\cline{2-3}
 & Size & Dim.  &   \\ \hline

SIFT1M~\cite{pq} & 1M  & 128 & SIFT~\cite{low04} \\ \hline
VLAD10M~\cite{boostkmeans} & 10M & 512 & VLAD~\cite{JPDSPS11} from YFCC~\cite{yfcc} \\ \hline
Glove1M~\cite{glove} & 1M & 100 & Vectorized text word~\cite{glove}\\ \hline
GIST1M~\cite{pq} & 1M & 960 & GIST~\cite{gist} \\ \hline
\end{tabular}
}
\label{tab:datasets}
\vspace{-0.2in}
\end{center}
\end{table}

\subsection{Evaluation Protocol}
Similar as~\cite{boostkmeans, ikmn15}, the average distortion (or mean squared error~\cite{pq}) is adopted to evaluate the clustering quality. Basically, it is the average distance between samples and their cluster centroid, which is given in Eqn.~\ref{eqn:distor}. As seen from the equation, it is nothing more than taking the average over $k$-means objective function Eqn.~\ref{eqn:tkm}. The lower the distortion value, the better quality of the clustering result. This measure is the same as within-cluster sum of squared distortions (WCSSD) in~\cite{cckm12}. 
\begin{equation}
\mathcal{E} = \frac{\sum_{q(x_i)=r}{\parallel C_r -x_i \parallel^2}}{n}.
\label{eqn:distor}
\end{equation}

In order to study the relation between the quality of KNN graph and the quality of clustering result, the average recall of KNN graph is also considered. In our evaluation, only the recall of top-1 nearest neighbor is measured. For SIFT1M dataset, the ground-truth of KNN graph is produced by brute-force search, which takes more than \textit{20} hours. While for VLAD10M dataset, it is too costly to produce the ground-truth for the whole set, the recall is therefore estimated by only considering nearest neighbors of \textit{100} randomly selected  samples.
 
\subsection{Configuration Test}
\label{sec:exnn}
In this section, different configurations on Alg.~\ref{alg:kmeans} are tested. As we discussed in Section~\ref{sec:alg}, Alg.~\ref{alg:kmeans} could be supported by other KNN graph construction algorithm. In addition, we also pointed out that similar speed-up scheme in Alg.~\ref{alg:kmeans} is feasible for traditional $k$-means. In this section, three different configurations are tested. In the first run, Alg.~\ref{alg:kmeans} is supplied with KNN graph from NN Descent~\cite{weidong}, which is denoted as ``KGraph+GK-means'' run. In the second run, Alg.~\ref{alg:kmeans} is modified to being built upon traditional $k$-means, which is denoted as ``GK-means$^-$'' run. In the standard setup run ``GK-means'', the clustering is built upon boost $k$-means. For both ``GK-means$^-$'' and ``GK-means'', the KNN graph is supplied by Alg.~\ref{alg:knng}. The experments are conducted on SIFT1M dataset. For all the runs, the cluster number is fixed to \textit{10,000}. 

\begin{figure}
\begin{center}
  \includegraphics[width=0.70\linewidth]{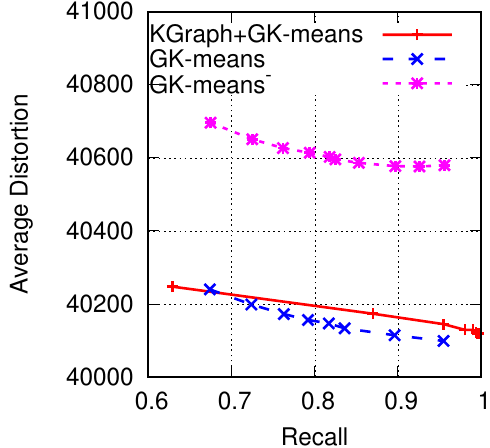}
  \caption{Configuration test on Alg.~\ref{alg:kmeans}. Alg.~\ref{alg:kmeans} is tested with the support of KGraph~\cite{weidong}. In addition, Alg.~\ref{alg:kmeans} is modified to doing clustering with traditional $k$-means.}
\label{fig:2knng}
\end{center}
\end{figure}
Fig.~\ref{fig:2knng} shows the distortion trend of these three configurations when KNN graphs of different qualities (reflected by their recall) are supplied. Basically, for all the configurations, higher KNN graph quality leads to steadily lower clustering distortion. 
When the KNN graphs are on the same recall level, GK-means built upon boost $k$-means shows much lower clustering distortion. Furthermore, GK-means converges to slightly lower distortion when KNN graph is supplied by Alg.~\ref{alg:knng}. Comparing with KNN graph supplied by NN Descent, KNN graph from Alg.~\ref{alg:knng} carries the information that how samples should be roughly organized as clusters since KNN list is built in Alg.~\ref{alg:knng} based on clustering results. In the following, due to its superior performance, the run ``GK-means'' is selected as the standard configuration of Alg.~\ref{alg:kmeans} for further comparison.

\subsection{Clustering Quality}
\begin{figure}
\begin{center}
	\subfigure[distortion vs. iteration]
  {\includegraphics[width=0.485\linewidth]{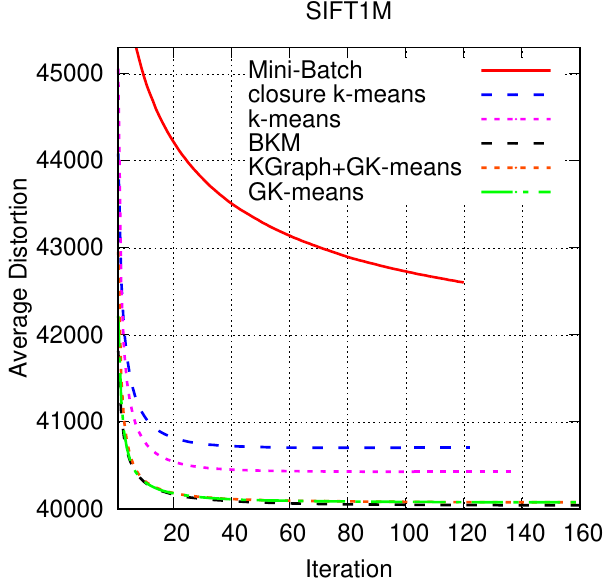}}
  \hspace{0.01in}
	\subfigure[distortion vs. time]
  {\includegraphics[width=0.485\linewidth]{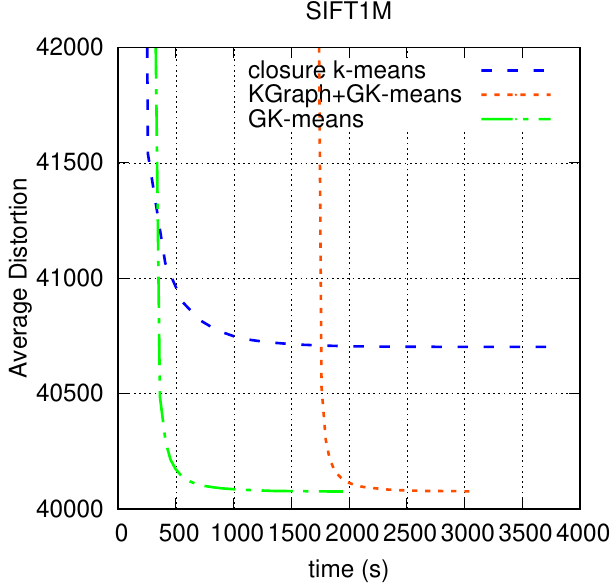}}
   \subfigure[distortion vs. iteration]
  {\includegraphics[width=0.47\linewidth]{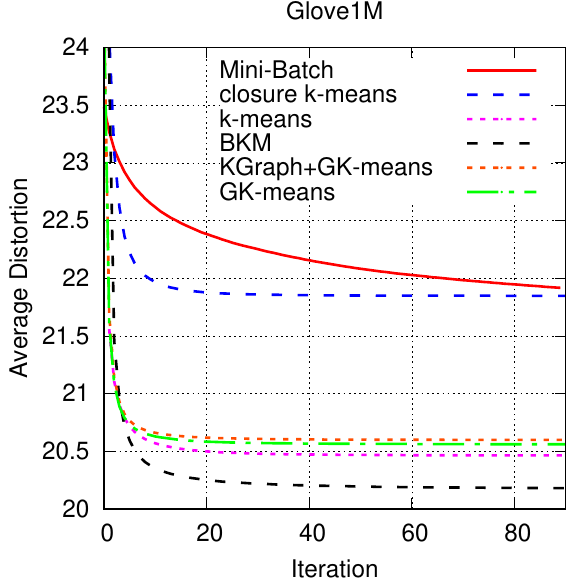}}
  \hspace{0.01in}
  \subfigure[distortion vs. time]
  {\includegraphics[width=0.49\linewidth]{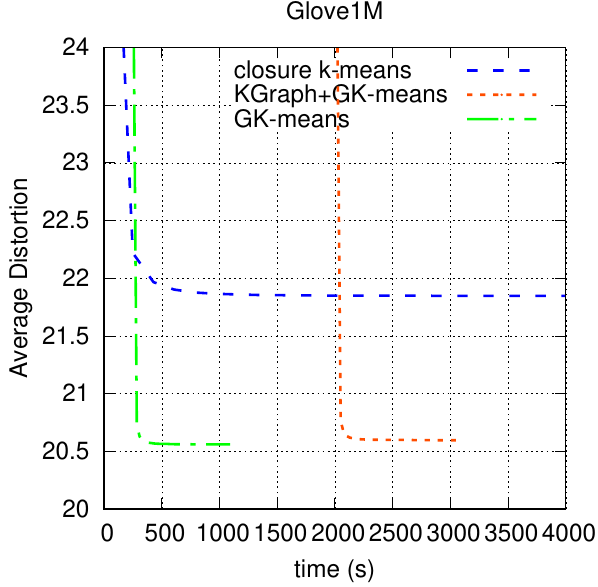}}
 \subfigure[distortion vs. iteration]  
  {\includegraphics[width=0.47\linewidth]{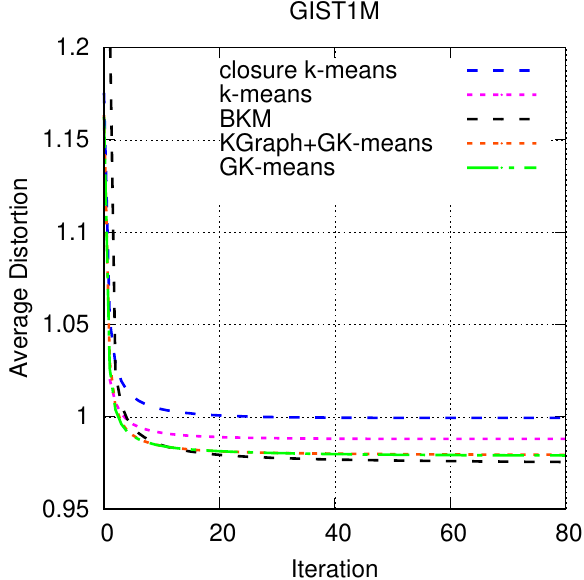}}
  \hspace{0.01in}
	\subfigure[distortion vs. time]
  {\includegraphics[width=0.49\linewidth]{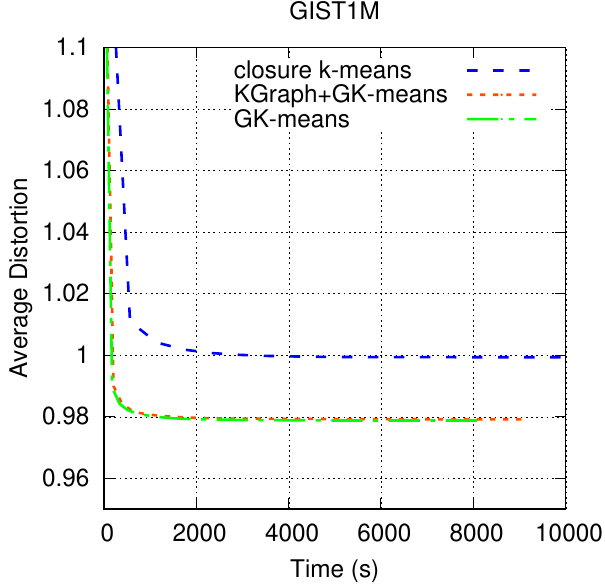}}
  \caption{Average distortion as a function of iteration times (shown in (a), (c) and (e)) and as a function of running time (shown in (b), (d) and (f)).}
\label{fig:distort}
\end{center}
\end{figure}

In this section, the clustering quality of GK-means is studied in comparison to $k$-means, boost $k$-means (BKM), closure $k$-means and Mini-Batch. The quality of fast $k$-means clustering is measured by studying the trend of clustering distortion against the number of iterations. Experiments are conducted on datasets SIFT1M, Glove1M and GIST1M. The cluster number is fixed to \textit{10,000} in all the experiments. Fig.~\ref{fig:distort}(a), (c) and (e) show the trend of clustering distortion as the function of iteration for datasets SIFT1M, Glove and GIST1M respectively. While Fig.~\ref{fig:distort}(b), (d) and (f) show the trend of clustering distortion as the function of clustering time for the algorithms that make a relatively good trade-off between efficiency and quality. The performance of \textit{k}-means, boost \textit{k}-means and Mini-Batch are not presented due to their efficiency or distortion (notably Mini-Batch) are not on the same level as GK-means and closure $k$-means.

As shown from Fig.~\ref{fig:distort}(a), (c) and (e), for all the methods except Mini-Batch, the clustering distortion changes very little after 30 iterations. Boost $k$-means always demonstrates the best performance in terms of clustering quality. In most of the cases, GK-means shows only slightly lower clustering quality than boost $k$-means. On SIFT1M and GIST1M, it even outperforms traditional $k$-means. KGraph+GK-means achieves similar performance as GK-means across all datasets. However, it is around 2 times slower since it is more costly to construct KNN graph by NN Descent. GK-means shows highest efficiency under all tests. Overall, GK-means offers a much better trade-off between efficiency and clustering quality among all the existing $k$-means variants.

\subsection{Scalability Test on Image Clustering}
\label{sec:tm}
In this section, the scalability of GK-means is tested on VLAD10M. In the test, the number of iterations for all $k$-means variants is fixed to \textit{30}. 

\begin{figure}[t]
\begin{center}
	\subfigure[time vs. size]
  {\includegraphics[width=0.485\linewidth]{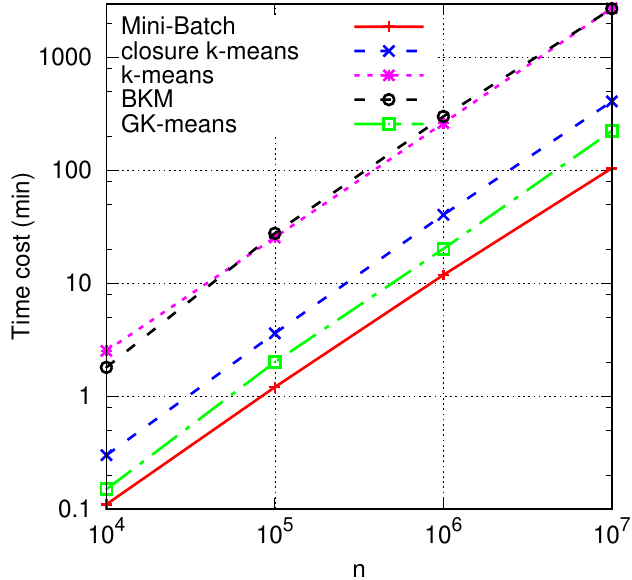}}  
	\subfigure[time vs. cluster num]
  {\includegraphics[width=0.485\linewidth]{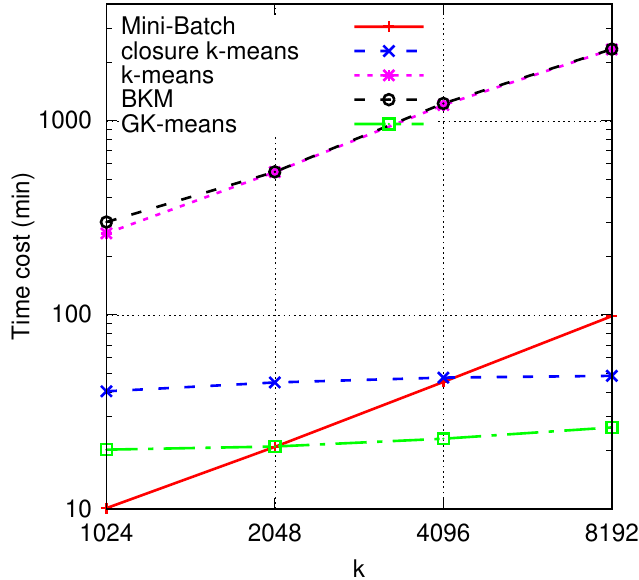}}  
	\caption{Scalability test on Flickr10M by varying the scale of input data: (a) and by varying the number of clusters: (b). }
	\label{fig:tm}
\end{center}
\end{figure}

\begin{figure}
\begin{center}
	\subfigure[k=\textit{1024}, varying \textit{n}]
  {\includegraphics[width=0.485\linewidth]{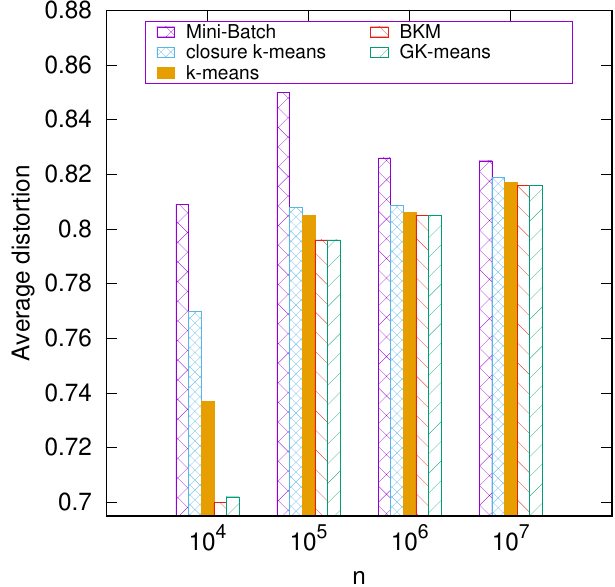}}
  \hspace{0.01in}
	\subfigure[n=$10^6$, varying \textit{k}]
  {\includegraphics[width=0.485\linewidth]{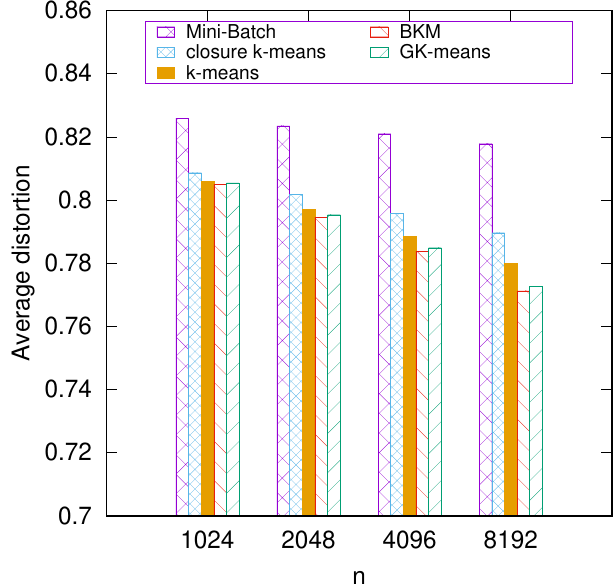}}  
	\caption{Average distortion from all \textit{5} methods under two different scalability testings on Flickr10M (best viewed in color).}
	\label{fig:dstall}
\end{center}
\vspace{-0.2in}
\end{figure}

In the first experiment, clustering methods are tested in the way that the scale of input images varies from 10K to 10M. For data in different scales, they are clustered into fixed number of clusters, i.e., \textit{1,024}. The time costs for all the methods are presented in Fig.~\ref{fig:tm}(a). Accordingly, the average distortion of all the methods are presented in Fig.~\ref{fig:dstall}(a).

As shown from Fig.~\ref{fig:tm}(a), GK-means is constantly faster than closure $k$-means and at least \textit{10} times faster than $k$-means and boost $k$-means. In the mean time, as shown in Fig.~\ref{fig:dstall}, clustering quality of GK-means is close to boost $k$-means across different scales of input data. In contrast, although Mini-Batch demonstrates fastest speed in this test, its clustering quality turns out to be very poor under different settings as is shown in Fig.~\ref{fig:dstall}(a).

%\begin{table}
%\begin{center}
%\caption{KNN Graph (top 50) construction in different size}
%\vspace{-0.1in}
%\footnotesize{
%\begin{tabular}{|l|c|c|c|c|}
%\hline
%\multirow{2}{*} {method} & \multicolumn{4}{c|}{time cost (minutes)}  \\ 
%
%\cline{2-5}
% & 10k & 100k & 1M & 10M   \\ \hline
%
%KGraph  & 1 & 12.3 & 133.6 & 1641   \\ \hline
%GK-means  & 0.1 & 0.7 & 14 & 173 \\ \hline
%\hline
%
%
%\multirow{2}{*} {method} & \multicolumn{4}{c|}{recall}  \\ 
%
%\cline{2-5}
% & 10k & 100k & 1M & 10M   \\ \hline
% 
%
%KGraph  & 0.98 & 0.95 & 0.88 & 0.40   \\ \hline
%GK-means  & 0.83 & 0.53 & 0.39 & 0.08 \\ \hline
%
%\end{tabular}
%}
%\label{tab:10m1m}
%\vspace{-0.2in}
%\end{center} 
%\end{table}

In addition, the scalability of clustering methods is tested in the way that the number of clusters varies from \textit{1,024} to \textit{8,192}, while the scale of input data is fixed to \textit{1} million. Fig.~\ref{fig:tm}(b) shows the time cost of all \textit{5} methods. Accordingly, the average distortion from these methods are given in Fig.~\ref{fig:dstall}(b). As shown in the figure, for $k$-means, boost $k$-means and Mini-Batch clustering methods, the time cost increases linearly as the number of clusters increases. Mini-Batch is no longer efficient as \textit{k} increases. In contrast, the time cost of closure $k$-means and GK-means remains nearly constant across different cluster numbers. In terms of clustering quality, as seen from Fig.~\ref{fig:dstall}(b), GK-means demonstrates similar quality as boost $k$-means and it is considerably better than closure $k$-means, Mini-Batch and $k$-means. A clear trend is observed from Fig.~\ref{fig:dstall}(b), methods based on boost $k$-means shows increasingly higher performance than the rest as \textit{k} grows. Overall, clustering driven by the proposed optimization process shows higher speed and better quality. The highest speed is achieved by GK-means, for which only \textit{18} minutes are required to cluster \textit{1} million \textit{512}-dimensional data into \textit{8,192} clusters.

Another more challenging scalability test is also conducted, in which VLAD10M is partitioned into \textit{1} million clusters. Two workable algorithms in such case, namely closure $k$-means and GK-means are tested. For GK-means, besides the standard configuration, GK-means that KNN graph is supplied by NN Descent is also tested, which is denoted as ``KGraph+GK-means''. Their performance is shown in Tab.~\ref{tab:10m1m}. For GK-means and KGraph+GK-means, the recall level of the approximate KNN graph is also reported. As shown in the table, compared to closure $k$-means, the runs from GK-means show significantly lower clustering distortion. In particular, GK-means with standard configuration shows the lowest clustering distortion. Similar as the experiments in Section 5.2, GK-means shows better performance when the KNN graph is supplied by Alg.~\ref{alg:knng}. KNN graph provided by Alg.~\ref{alg:knng} keeps the information of intermediate clustering structures. Such kind of information will be  transferred to the clustering process. It is therefore able to produce better quality even though the recall of its KNN graph is lower than that of NN Descent. GK-means also achieves the highest speed efficiency in such a challenging test. According to our estimation, it would take more than \textit{3} years to fulfill the same task for traditional $k$-means. 

\begin{table}
\begin{center}
\caption{Performance of GK-means and closuer $k$-means when partitioning VLAD10M into 1M clusters. The time costs in clustering initialization and $k$-means iterations are shown}
\vspace{-0.1in}
\footnotesize{
\begin{tabular}{|l|c|c||c|c|c|c|}
\hline
\multirow{2}{*} {Method} & \multicolumn{3}{c|}{Time cost (h)} & \multirow{2}{*} {$\mathcal{E}$} & \multirow{2}{*} {Recall} \\ 
\cline{2-4}
 & Init. & Iter. & Total & &  \\ \hline
%NNG50 1024 & 2.6 & 0.6 & 3.2 & 0.817269\\ \hline
%Closure 1024 & 0.9 & 5.9 & 6.8 & 0.818985\\ \hline
KGraph+GK-means & 27.3  & 3.2 & 30.5 & 0.649 & 0.40 \\ \hline
%ITK (25) & 2.5 & 1.3 & \textbf{3.8} & 0.622& 0.08 \\ \hline
GK-means & 2.7 & 2.5 & \textbf{5.2} & \textbf{0.619}& 0.08 \\ \hline
Closure $k$-means & 0.9 & 9.6 & 10.5 & 0.700& N.A. \\ \hline
\end{tabular}
}
\label{tab:10m1m}
\vspace{-0.2in}
\end{center} 
\end{table}

\section{Conclusion}
\label{sec:conl}
In this paper, we have presented our solution to the scalability issue of $k$-means. We show that fast $k$-means clustering is achievable with the support of an approximate KNN graph. Specifically, in the $k$-means iteration, one sample only needs to compare with clusters that its nearest neighbors live in. The clustering complexity is therefore irrelevant to clustering number. As shown in the paper, hundreds to thousands times speed-up is achieved in particular in the case that both \textit{n} and \textit{k} are very large. In addition, since the fast $k$-means is built upon boost $k$-means, it also shows very high clustering quality. Overall, the proposed GK-means shows considerably better trade-off between clustering quality and efficiency over existing solutions. Moreover, the beauty of this algorithm also lies in the design of fast KNN graph construction process. The KNN graph is built by calling GK-means itself in an intertwined evolving process. In the process, the KNN graph and $k$-means clustering are incrementally optimized. This intertwined self-evolving process could be generalized as an unsupervised learning framework, which will be our future research work.

\section*{Acknowledgement}
This work is supported by National Natural Science Foundation of China under grants 61572408.

\bibliographystyle{ieeetr}  
\bibliography{chenghaod}

\end{document}